# EEG-based Drowsiness Estimation for Driving Safety Using Deep Q-Learning

Yurui Ming, Dongrui Wu, Yu-Kai Wang, Yuhui Shi, Chin-Teng Lin

*Abstract*— **Fatigue is the most vital factor of road fatalities, and one manifestation of fatigue during driving is drowsiness. In this paper, we propose using deep Q-learning to study the correlation between drowsiness and driving performance. This study is carried out by analyzing an electroencephalogram (EEG) dataset captured during a simulated endurance driving test. Driving safety research using EEG data represents an important brain-computer interface (BCI) paradigm from an application perspective. To formulate the drowsiness estimation problem as an optimization of a Q-learning task, we adapt the terminologies in the driving test to fit the reinforcement learning framework. Based on that, a deep Q-network (DQN) is tailored by referring to the latest DQN technologies. The designed network merits the characteristics of the EEG data and can generate actions to indirectly estimate drowsiness. The results show that the trained model can trace the variations of mind state in a satisfactory way against the testing EEG data, which confirms the feasibility and practicability of this new computation paradigm. By comparison, it also reveals that our method outperforms the supervised learning counterpart and is superior for real applications. To the best of our knowledge, we are the first to introduce the deep reinforcement learning method to this BCI scenario, and our method can potentially be generalized to other BCI cases.**

## I. Introduction

FATIGUE is regarded as the most severe factor causing road fatalities [1]. To understand the correlation between fatigue and driving performance, both from theory to practice, is of persistent interest for researchers. There have already been many attempts and substantial effort spent in understanding the impact of fatigue – such as drowsiness on drivers' response [2-4]. Generally, physiological indicators that distinguish drowsiness from the normal state, can facilitate the research with the aim to improve road safety. Previous work based on bio-traits include but not limited to eye blink [5, 6], heart rate [7], respiration [8], facial expression [9]. In addition, the technology advance also promotes the usage of portable devices to study driving fatigue [10, 11]. Among the research in recent years, electroencephalogram (EEG) which is the direct measurement of brain activity indicating the mind state, gains more emphasis and popularity [12-16]. Furthermore, in light of the trend that dry sensors are maturing to the stage of high-quality EEG signal acquisition, research via EEG is one promising approach [17, 18]. By monitoring the mind state via a potential portable EEG device, the brain can directly communicate with the driving environment in a spontaneous and passive manner to ensure driving safety.

However, due to the characteristics of brain structures, direct measurement of the mind state, such as drowsy or alert, is extremely difficult. Hence, the study of driving safety is usually carried out in a laboratory environment via some approaches ingeniously designed to indirectly measure the mind state [13]. Based on measured performance indicators such as the response time (RT) [13], some models have been established to analyze the signals and deployed for application at a later stage.

Some conclusions have already been reached using traditional machine learning methods especially from the perspective of supervised learning [19-21]. Moreover, due to the wide spectrum of achievements via deep learning (DL), there have also been a variety of attempts to adopt DL for drowsiness detection via EEG with good results [22-25]. However, the unique characteristics of EEG data still incur limitations by using supervised learning that affect the theoretical analysis and real applications. The first problem is the efficiency of data utilization. Supervised learning models, including deep neural networks (DNNs), require the regularized input data to be paired with ground-truth labels for training and testing. If there is no label information, the corresponding EEG segments will be unavoidably discarded. Second, some models that require measured information such as RT for performance tuning only work in an offline manner and are unsuitable or nonapplicable for online use, especially from an engineering perspective.

In this paper, we introduce deep reinforcement learning, specifically deep Q-learning, to analyze EEG data captured in an experiment of driving safety research. There are research works jointly considering EEG and (deep) reinforcement learning; however, these researches are based on the reward prediction error theory of dopamine [26-28], and mainly use the EEG signal as the error signal, i.e., the reward [29-31]. One eminent characteristic of reinforcement learning is that the future reward can be used to assess the current action to take given the current state. In our work, a function involving the measured RT in some latency yields the reward, which is used to assess the action against the current state (EEG data segment). Another aspect is that for reinforcement learning, although the training phase requires that costly label information (RT) be obtained for reward calculation, for the testing stage, an optimal policy is always assumed and exploited for action selection. Thus, no further deliberate articulated measured information is needed. These facts suggest a promising potential for analyzing the EEG signals of similar BCI scenarios for real applications. To the best of our knowledge, we are the first to assess driving safety via EEG using a reinforcement learning method.

The rest of this paper is structured as follows. In section II, we describe the problem and the simulated experiment in detail to formulate the corresponding reinforcement learning model. Section III details a deep Q-network that is tailored for analyzing the EEG data captured from a simulated driving safety test. In section IV, by comparing the performance of our method to the corresponding supervised learning model, we demonstrate the practicability of our proposed approach. Section V discusses the research background and interesting discovery of using this method from a neurophysiological perspective.

## II. PROBLEM DESCRIPTION

### A. Driving Safety Research

The investigation of causality between fatigue and driving performance requires the driver's implicit mind states and corresponding reactions. It is more practical to first carry out the research in the laboratory to simulate the problem, to explore potential solutions, and to assess the proposed method, etc., for later generalization to practical applications. Thus, a simulated environment is built to mimic the endurance driving scenario for subsequent research.

The experiment is designed as recruited drivers performing endurance driving along the simulated highway and generally lasts 90 minutes. It is postulated that during the experiment, the subjects' physiological state is hardly maintained at the same condition. The experiment was approved by the Institutional Review Board of the Taipei Veterans General Hospital (PN: 101W963, VGHIRB No.: 2012-08-019BCY). The voluntary, fully informed consent template for subjects was reviewed by the ethical committee (IRB -TPEVGH SOP 05). The consent was signed by all the subjects who participated in this research, as required the by accompanying regulations. Fig. 1 (A) illustrates the driving scenario on a four-lane highway. The horizon for the driver is projected on a large chained screen. To simulate different driving conditions on the road, a real car is converted and installed on a maneuverable platform with six-degree freedom to guarantee sufficient flexibility, as shown in Fig. 1 (B).

The procedures of the experiment are as follows. The driver operates the car cruising in one lane of the highway. Random turbulence is deliberately introduced to cause the car to deviate from the original cruising lane (event onset). The driver is instructed on observation to quickly turn the steering wheel (response onset) for compensation to move the car back to the original lane. Once the car returns to its original cruising state (response offset), the trial is complete. Notably, fatigue can be attributed to many factors. In this work, our consideration is mainly from the drowsy perspective. To correlate drowsiness with driving performance, the driver only needs to operate the steering wheel and is free from brake and accelerator controlling.

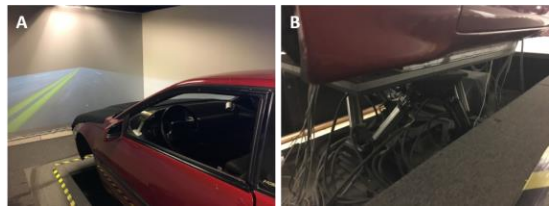

Fig. 1 Simulated driving setup: (A) the mimicking highway scenario; (B) the maneuvering platform with the installed converted car.

EEG signals are continuously captured during the whole experiment via the Scan SynAmps2 Express system [32]. A wired EEG cap with 32 Ag/AgCl electrodes is used, including 30 EEG electrodes and 2 reference electrodes (opposite lateral mastoids). The EEG montage is in accordance with a modified international 10-20 system [33, 34]. The contact impedance between electrodes and the scalp is kept below 5kΩ. The EEG recordings are digitized at 500 Hz with 16-bit quantization.

Because fatigue cannot be directly measured, an indirect indicator, RT, is introduced for this purpose. In detail, the duration between the event onset to the response onset is termed RT, as aforementioned. It measures how quickly the subject reacts to a stimulus. The period before event onset is called the baseline region. Depending on the configuration of the trial interval, the duration of baseline can vary but lasts for at least 5 seconds (s). EEG data from this region are nominated for brain dynamic analysis and machine learning algorithms. We also illustrate the corresponding terminologies that are key to understanding the experiment in Fig. 2.

There are 37 subjects participated in the experiments, with an average age of 22.4. Most of them are undergraduate students without previous experiencing in EEG. The experiment can be easily configured via device settings, such as different EEG montages, arousal signal emission, etc., to investigate different aspects during driving. We only consider the EEG data captured based on the setup (device) configuration mentioned above. Usually, the number of times a subject participates in the experiment is not mandatory in order to encourage voluntary dedication, although all subjects are paid. To ensure data quality, we host only one subject doing the experiment per day. The EEG data captured during one participation of any subject in the experiment are called session data. The number of sessions for each subject depends on the number of his/her participations. Each session data contains numerous trial data. The interval between consecutive trials is 5 to 10 s at random. The number of trials in one session is highly dependent on the situation when subjects participated in the experiment. For example, if the subject mostly experiences sleepiness during the experiment, the bulk of trials could last longer due to poor reactions. Because the whole process is always approximately 90 minutes, variations in RT can induce different numbers of trails of separate sessions.

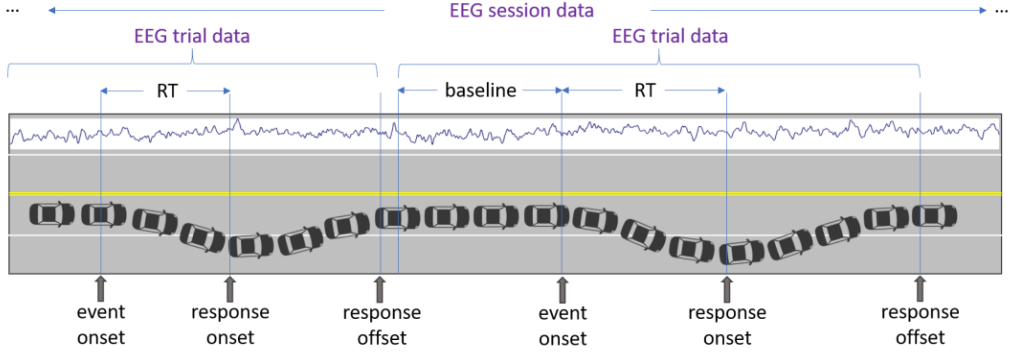

Fig. 2 Illustration of the experimental scenario. A trial launches from the event onset and extends to the response offset. A session is constituted by repetitive trials.

Notably, although there lacks an explicit formula linking RT with the mind state, it is postulated that they are positively correlated. The correlation can be estimated from the corresponding EEG data captured during the experiment. By introducing a reasonable threshold, RT is sufficient from the application perspective. For example, any RT value above the threshold can be regarded as poor control of the car, and a warning should be raised to prevent potential accidents.

*B. Reinforcement Learning Proposition*

Inspired by human's goal-directed learning process, reinforcement learning is modeled as an agent interacts with the environment $\mathcal{E}$ by balancing between the exploration and exploitation for a maximum return $G$ (the accumulation of reward $r_t$) in the long run. The agent's decision regarding the action $a_t$ is based on the current state $s_t$ of the environment by referring to the policy $\pi$. This sequential decision-making process is required in accordance with the Markov decision process (MDP) for mathematical rigour, which backs the convergence of the general iterative optimization process [35]. To evaluate different policies, the state function $V$ and value function $Q$ are defined as follows:

$$Q_\pi(s, a) = \mathbb{E}_\pi[R_t | S_t = s, A_t = s] \quad (1)$$

$$V_\pi(s) = \mathbb{E}_{a \sim \pi(s)}[Q_\pi(s, a)] \quad (2)$$

$$Q_\pi(s, a) = \sum_{s', a'} p(s'|s, a)\pi(a|s)[r + Q_\pi(s', a')] \quad (3)$$

One step induction of (1) leads to the formula (3), which obeys the Bellman equations [35]. Determining the Q value of each pair $(s, a)$ for each policy $\pi$ is not the target; the purpose is to find the optimal policy $\pi^*$ satisfying (4) and (5) by policy iteration:

$$Q_{\pi^*}(s, a) = \max_\pi Q_\pi(s, a) \quad (4)$$

$$V_{\pi^*}(s) = \max_\pi V_\pi(s) \quad (5)$$

There are several methods to achieve this goal; however, for brevity, we directly turn to Q-learning, which is an off-policy temporal difference algorithm [35]:

$$Q(S_t, A_t) \leftarrow Q(S_t, A_t) + $$
$$\alpha \left[ R_{t+1} + \gamma \max_a Q(S_{t+1}, a) - Q(S_t, A_t) \right] \quad (6)$$

The intuition here is that the learned $Q$-function directly approximates the optimal function $Q_{\pi^*}$ independent of the policy being followed. Notably, conventional Q-learning is capable of solving problems only with limited state space and action choice. For complex problems with large state space, one idea is to map the state space to action preference using a function. This is the work systematically investigated by DeepMind [36-39]. Optimization in this way is specifically governed by the following formula (7):

$$L_t(\theta_t) = \mathbb{E}_{s, a \sim \rho(\cdot)} \left[ \left( y_t - Q(s, a; \theta_t) \right)^2 \right] \quad (7)$$

$$y_t = \mathbb{E}_{s' \sim \mathcal{E}} \left[ r + \gamma \max_{a'} Q(s', a'; \theta_{t-1}) \right] \quad (8)$$

where $Q$ is the action-value function parameterised by $\theta$. $\rho(\cdot)$ indicates the trajectory distribution, and $\mathcal{E}$ indicates the state distribution in a certain environment. We will readdress DeepMind's contribution in optimizing (7) later.

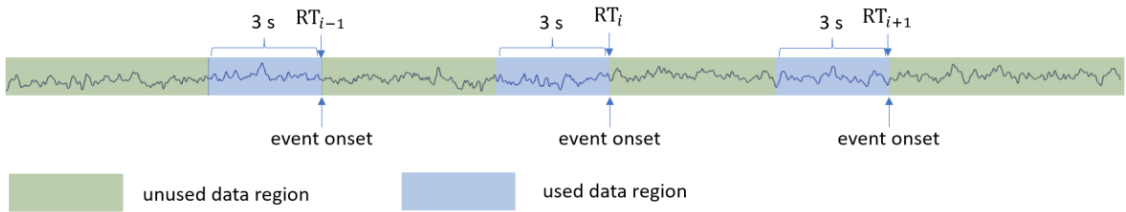

Fig. 3 Data utilisation paradigm of supervised learning. Only EEG data under the blue mask are used for training and validation.

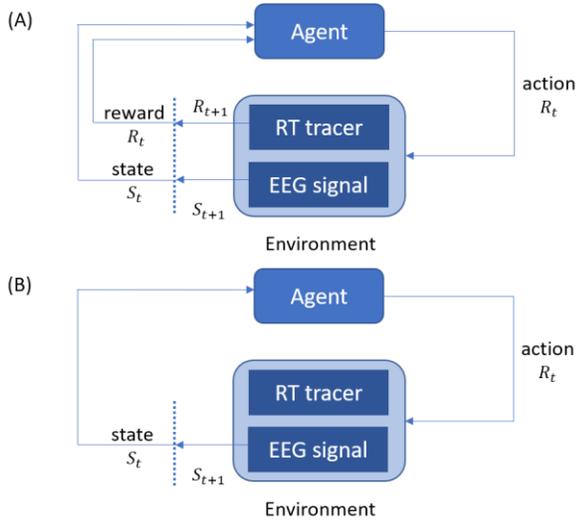

Fig. 4 Reinforcement learning paradigm for RT prediction (A) training stage; (B) testing stage.

To adjust the driving safety problem resulting in a solution from reinforcement learning algorithms, a retrospect of supervised learning is presented to highlight the challenges. For supervised learning, the EEG data in the baseline region are extracted and coupled with the measured RT for training and testing. Exemplified by the work in [13, 22], in Fig. 3, only EEG signals covered by the blue masks are utilized during the analysis, because it is ideal to extract brain dynamics there [14]. Most EEG data covered by green masks are discarded due to the lack of label or RT information, which indicates insufficient data utilization. However, when deploying the trained model for application, the captured EEG data are continuously input for inference about the mind state. There is no differentiation of the EEG data segments from the temporal perspective. Such a wider generalization scope can incur poor performance due to the discrepancy in the distributions between the EEG data for training and for application.

The paradigm of reinforcement learning requires abstraction and instantiation of the agent, environment, state, action and reward to target a specific problem, which is assumed to comply with MDP. The precaution for driving safety can be alternatively attributed to the good prediction of RT, based on which appropriate actions can be taken. The criterion for the quality of predicted RT is its deviation from the corresponding measured RT. When comparing the two, the less the deviation, the better the quality. Therefore, we need a mechanism to maintain the predicted RT for comparison with the measured RT when necessary (i.e., where there exists a measured RT for the current state). To summarize, the environment can be designed as comprising the captured EEG data and an internal RT tracer. The agent issues actions to manipulate the tracer to estimate RT. This strategy leads to the design of a reinforcement learning architecture shown in Fig. 4.

An instantiation of the reinforcement learning method based on the above depiction is illustrated in Fig. 5. The components of reinforcement learning are dissected into a concrete representation in the context of session data captured in one experiment. As shown in Fig. 5, the state $s_t$ is one piece of the continuously segmented session data, and the action $a_t$ indicates how to operate the internal tracer. To fabricate a reward at every time step, the following strategy is taken: if the duration of state $s_t$ covers a measured RT, $r_t$ is equal to the negative of the absolute difference between the measured RT (mRT) and the traced RT (tRT); otherwise, the reward $r_t$ is equal to 0. Note that we take the negative of the absolute value because the optimization targets a maximum return. Since we focus on Q-learning, the operation to the internal tracer is specially designed to allow only discrete operations, which is intrinsically required by the algorithm itself [35]. As shown in Fig. 5, an exemplified action can be chosen to maintain the current traced RT or to increase/decrease the RT by a certain unit such as 0.5 s. Based on the above description, the interaction between the agent and environment is as follows. At time step $t$, by analyzing the EEG data segment $s_t$, the agent acts according to $a_t$ to operate tracer with a consequent reward $r_t$. Then the traced value is updated which is transparent to the agent, and environment evolves into the new state $s_{t+1}$.

It is pointed out here that although the actions designed above are self-evident from the illustrative perspective, upon implementation, we use a mechanism called RT proposition. The problem for the action paradigm in Fig. 5 is that the final predicted RT by tracing is history-dependent. Although it is a more general case than Markov-dependent, history-dependence might not be flexible enough. Meantime, it can be roughly postulated that the mind state at $t + 1$ only depends on its predecessor at $t$. Our empirical experience can justify this. Supposing that we are sleepy at time step $t$, we can either feel sleepy or experience sudden alertness at $t + 1$. Either way, it is not necessary to check the mind state at $t - 1$. This argument leads to the formulation of the up-to-date traced RT value as follows:

$$\text{tRT}_t = \beta \cdot \text{tRT}_{t-1} + (1 - \beta) \cdot \text{pRT}_t \qquad (9)$$

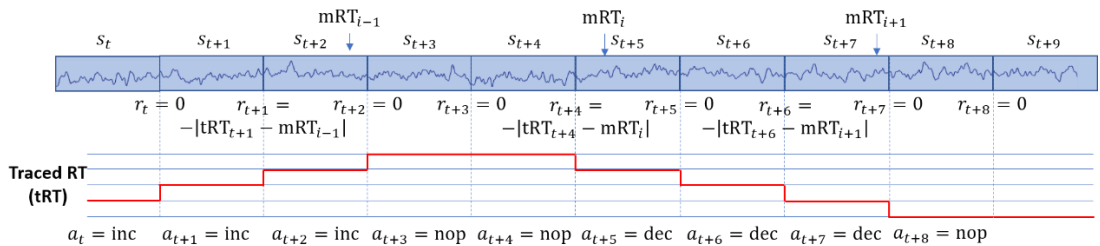

Fig. 5 Instantiation of concepts in reinforcement learning in the context of driving safety research.

In (9), pRT denotes the proposed RT at the current time step $t$ and can only have discrete values, such as 1.2 s, 2.5 s, etc., due to Q-learning being considered in this paper, as mentioned above. $\beta$ denotes the tradeoff between previous tRT and current pRT and is named transition weight. Because the traced RT is termed from the illustrative dynamic decision-making perspective, we use it interchangeably with the term predicted RT, which is more commonly used in supervised learning. We also note here that the predicted RT is different from the proposed RT, as the latter is only internal to the agent. The pRT bid by the agent via different actions is used to modulate the tRT, which is equivalent to the predicted RT unless otherwise specified.

Notably, compared with supervised learning, which focuses on trial data, reinforcement learning in our design works directly with session data. To the best of our knowledge, we are the first to introduce reinforcement learning in this field. It is mentioned that the currently available dataset might pose a challenge for our proposed method, because session data are not abundant due to the original experimental design. However, as long as the practicability can be demonstrated via the current dataset, it still indicates a promising beginning for subsequent research. We also point out that since our approach is from the prototyping perspective, it might lack of the field-engineering rigor. For example, the state in our design is the non-overlapping EEG segment in 3 s. The treatment is just derived from laboratory convention and might not be the optimal choice in real application.

## III. NETWORK DESIGN

DeepMind has performed revolutionary work in revising conventional reinforcement learning methods, such as Q-Learning. Engineers working there devised deep Q-learning, or DQN if the underlying DNN for the Q-function estimation needs to be emphasized [37], which was originally thought to be impractical due to several constraints. They were the first to introduce several tricks such as target network, experience replay, etc., which have paved the way for successfully applying DQN in different fields. They also enhanced DQN by incorporating more techniques from traditional reinforcement learning and proposing new architectures such as double DQN and duelling DQN [38, 39]. Our work adopts the same tricks during the utilization of DQN and makes some modifications to the experience replay part to achieve more efficient computation. As shown in Fig. 6, we add a batch buffer that actually refers to the frames in the replay queue to avoid unnecessary copying operations. We add a sequence number, which in theory can wrap back to 0 for each frame, to adjust the sample selection.

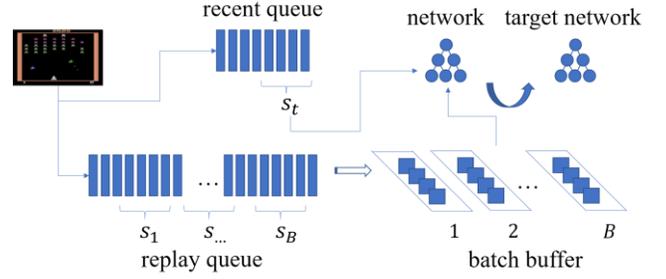

Fig. 6 Modification of experience replay.

To design a DQN that is suitable for analyzing EEG data for action or RT proposition, previous work in [40] is referenced to guide overall architecture. The principles include fewer pre-processing of input data to relieve the EEG domain knowledge requirement, effective computation to facilitate online deployment, and similar input-output interface to recycle learning algorithms with DQN and its variants. These considerations lead to the designed Q-network architecture shown in Fig. 7.

We elaborate on our proposed network architecture to a certain degree to ease the understanding of some technical details. First, the structure up to the RNN part (inclusively) is identical to the recurrent convolutional neural network (RCNN) structure as in [41]. It utilizes a weight-sharing CNN to extract spatial features for subsequent processing by RNN to further explore the temporal information. Accordingly, the input to the network are pre-processed multi-channel EEG data. As mentioned above, the session data are segmented into successive chunks in a certain unit, which is 3 s in this paper. For step $t$, the data $s_t$ which is sliced into three pieces of length 1 s, i.e., $s_t^1$, $s_t^2$ and $s_t^3$, are consecutively fed into the network for feature extraction. These convolutional operations constituting the CNN are arranged in a dashed box to indicate that the weights are shared between processing $s_t^1$, $s_t^2$ and $s_t^3$. The details of the variants of the standard convolutional operation, such as depthwise and separable convolution, can be found in [42]. After extracting relevant features via CNN, these features are stacked together to feed into the RNN for further processing.

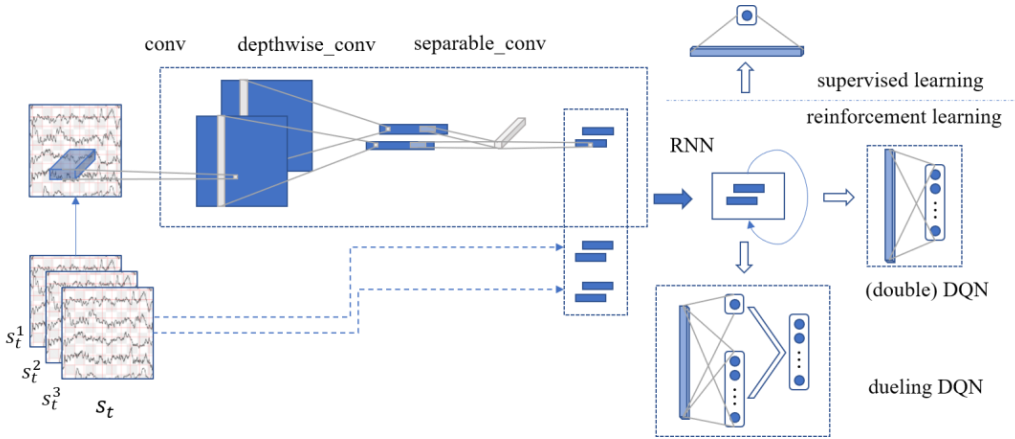

Fig. 7 Network architecture. Note that up to RNN, these components can be shared by both the supervised learning and reinforcement learning paradigms. For reinforcement learning, DQN or double DQN can share the structure up to RNN with dueling DQN.

TABLE I NETWORK ARCHITECTURE CONFIGURATION

| Model | Layer ID | Layer Type | Filters | Size | Act. | Pad. | Comment |
|---|---|---|---|---|---|---|---|
| CNN | 1 | Conv2D | 32 | (1, 64) | - | Same | |
| | 2 | Depthwise-Conv2D | 1 | (30, 1) | tanh | Valid | Weight clipping |
| | | AvgPool2D | - | (2, 2) | | Same | |
| | 3 | Separable-Conv2D | 32 | (1, 16) | tanh | Same | |
| | | AvgPool2D | | (2, 2) | | Same | |
| RNN | 4 | Conv2D | 4*3*32 | (1, 8) | - | Same | 2D LSTM |
| Supervised | | Linear | - | 512 | id | | $L_2$ Regularization |
| | | Linear | | 1 | | | |
| (Double) DQN | | Linear | - | 512 | id | | $L_2$ Regularization |
| | | Linear | | #actions | | | |
| Dueling DQN | state | Linear | - | 512 | id | | $L_2$ Regularization |
| | | Linear | | 1 | | | |
| | state-action | Linear | - | 512 | id | | $L_2$ Regularization |
| | | Linear | | #actions | | | |

Then, the dataflow from the RNN is conducted to multiple network components for different purposes, as depicted in Fig. 7. The subnetwork for supervised learning is commonly found in literatures and is hence self-explanatory [43]. By adding a fully connected layer with one neuron, the network in this circuitry is used as a regressor to predict RT.

For the (double) DQN in this work, it adopts a similar treatment as in [37], which is essentially by employing DNN as Q-function to boost the performance of Q-learning. In brief, suppose $s \in \mathcal{S} \subset \mathbb{R}^n$ and $a \in \mathcal{A} = \{1, 2, \cdots, |\mathcal{A}|\}$, the constructed network is not in line with the Q-function definition that maps $[\mathbb{R}^n, \mathcal{A}]$ to $\mathbb{R}$, but instead maps $\mathbb{R}^n$ to $\mathbb{R}^{|\mathcal{A}|}$ from the implementing perspective. Therefore, the DQN is constructed as a two-layered perceptron: one layer is used to adjust the dimension, and a final layer generates the Q value for each action. Further computations are based on the output from the network, such as employing the max operator to select the action to interact with the environment.

For dueling DQN, the insight lies in the observation that to estimate the value of each action for some states is unnecessary. So, two sub-networks are constructed to separately estimate value function $V$ and advantage function $A$ in accordance with (10):

$$Q(s, a,; \theta, \alpha, \beta) = V(s, a; \theta, \alpha) + A(s, a; \theta, \beta) \quad (10)$$

The definition of $A$, the advantage function that relates $V$ and $Q$ is as in (11):

$$A_\pi(s, a) = Q_\pi(s, a) - V_\pi(s) \quad (11)$$

$V$, $Q$ and $A$ in (10) are parameterized version of (1), (2) and (11), respectively, and they share part of the network, as reflected by the parameter $\theta$. In this paper, two streams concerning $V$ and $A$ are both implemented as MLP. As in Fig. 7, the branch with one neuron is used to estimate $V$, and the other branch is for $A$. The two estimates are added together as $Q$, which bears later stage computation for the functionality of the whole duelling network. The final network configurations for different models are detailed in TABLE I.

With the designed Q-network, the procedures for EEG data analysis can be summarized as the following steps:

- Carry out the experiment to obtain session data.
- Establish the measured information for reward calculation.
- Setup parameters such as segment length, and input the data into the model for training.
- Test the performance of the model for application.

## IV. EXPERIMENT

We use the EEG data captured during the simulated driving experiment described above to evaluate our proposed method. One trait of EEG data that tends to affect the performance is the intra- and cross-subject variance. For the experiment, we separately consider the single-subject and the multi-subject cases in the following sections. Because we are the first to employ DQN for driving safety research via EEG especially from an application perspective, the discrete action space requires limited RT resolution (due to discrete actions) to approach the measured RT, which in theory is a real number. This restriction causes difficulty in competing with the previous results achieved by very deep network structures especially due to quantization error. In addition, our pre-processing of the EEG data is rather lightweight; hence, our main consideration here is not to seek the best network structure to compete with the state-of-the-art result. Instead, by comparing with a supervised learning model that shares the network structure shown in Fig. 7, we would rather demonstrate the feasibility and practicability of this new application-oriented computation paradigm for driving safety research via EEG.

EEG signals are in general complicated due to signal distortion, artifacts pending, etc. In this work, only limited pre-processing is applied to the data and measured RT. For EEG signals, a bandpass filter with a range 0.5 to 50 Hz is introduced, followed by down-sampling to 128 Hz from the original sampling rate of 500 Hz. For RT, the values are clipped into the range 0.5 to 8 s, with a moving average within 90 s [44]. Finally, the post pre-processed continuous signals are segmented into consecutive sections with a unit of 3 s, as shown in Fig. 5.

*A. Single-subject Case*

Generally, the variance of the statistical distributions of the EEG data of a single subject is not as severe as that of multiple subjects. In light of this fact, we first consider the single-subject case to verify the performance of the proposed model. We choose subjects of the same gender that participated in the experiment at least three times. For some sessions, the experiment was configured to issue warnings upon event onsets, to inspect the influence of explicit arousal on driving performance. We omit these sessions for consistent experimental scenarios. The subjects that merit the above criteria are S01(4)(1410), S09(3)(704) and S41(3)(1405). The number in the first parentheses indicates the number of session data, i.e., the number of times that the subject participated in the experiment. The number in the second parentheses indicates total trials of all sessions. For each subject, the data from the first two sessions are used for training and the remaining session is for testing. One session data of S01 is used for validation to decide the total number of iterations which is applied for training on all subjects' data.

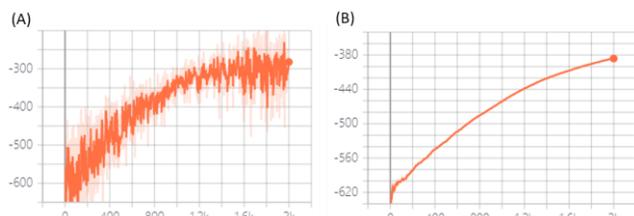

Fig. 8 Convergence indicators: (A) the episode return (B) the average return. The $x$-axis indicates the sequence of training iteration or episode. The $y$-axis indicates the return per episode.

We take the dueling DQN architecture due to its performance and train the model for 2000 iterations or episodes with similar hyperparameters adopted from [37]. The transition weight $\beta$ is set to 0.75. To monitor the convergence of the training process of our proposed model, two indicators are employed for inspection: the episode return and the average return. The episode return is the summation of rewards at each step when processing session data. It is a direct measurement of episodic gain but tends to be noisy. By introducing the average return over episodes, the trend for convergence is relatively clearer. However, due to the accumulation effect, there might be a latency of reflection for overfitting. Nevertheless, by combining the two indicators, the training process can be well monitored. We randomly switch between the data of two sessions during the training process, and Fig. 8 illustrates the corresponding episode return and average return. The trained model is applied

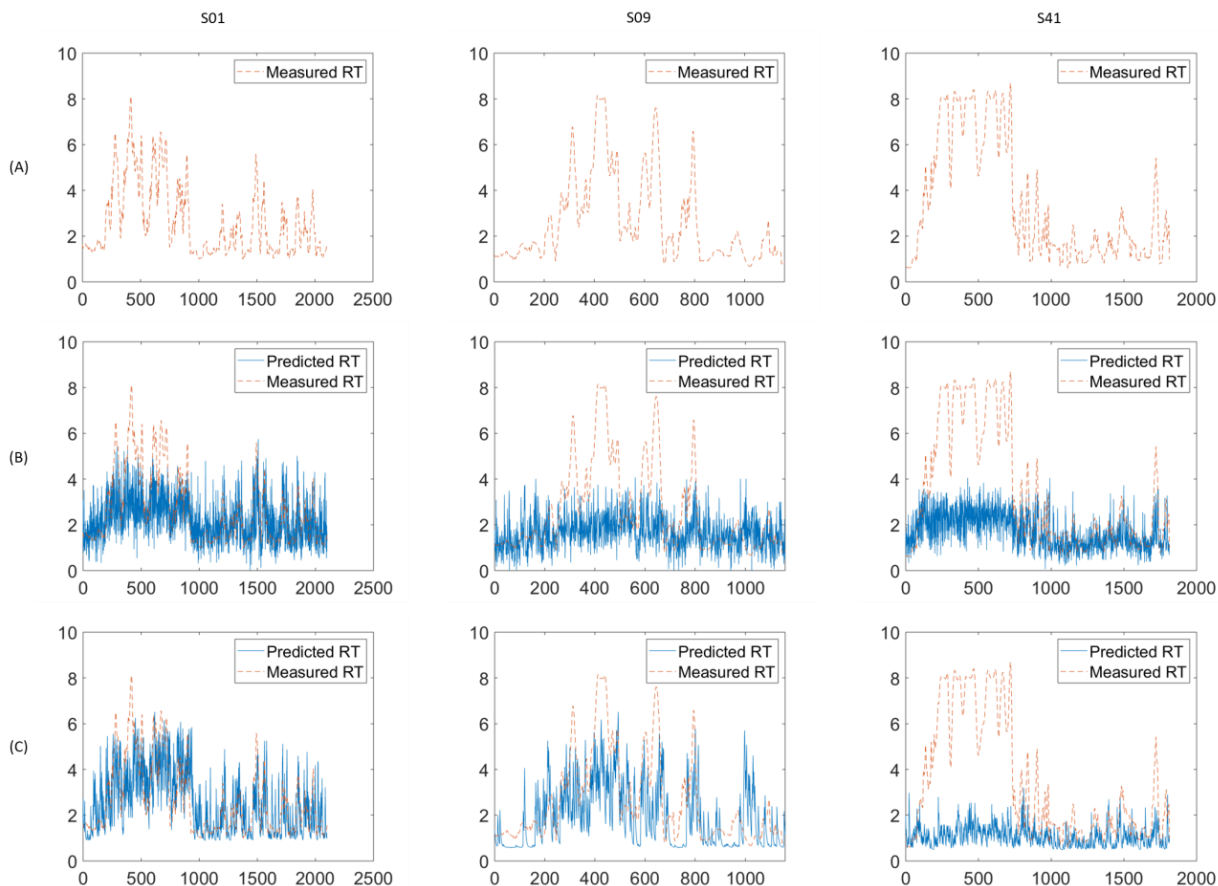

Fig. 9 Measured RT vs predicted RT for single-subject test: (A) measured RT; (B) supervised learning case; (C) reinforcement learning case. The $x$-axis indicates the progress of the session (3 seconds per step). The $y$-axis indicates the lengh of RT (unit: seconds).

to the test session data to predict RT in a consecutive manner.

For the benchmark method, we use the same network structure to construct the regressor and train in the supervised learning manner, according to Fig. 7 and TABLE I. The pre-processing of the EEG data is the same as the procedures for reinforcement learning. For each trial, 3 s of EEG signals in the baseline region are extracted, as in Fig. 3. Coupled with the measured RT, these input/label pairs are used for training and testing. To be consistent with the data division for training and testing as in the case of reinforcement learning, trial data of the first two sessions are used for training. The trial data of one session of S01 are for validation to decide the total number of iterations during training. The model is trained with the learning rate of 0.0001 for 600 iterations for each subject. The trained model is tested on trial data to predict the RT and also made inference on the test session in a consecutive manner, as in the case of reinforcement learning.

To quantitively assess the performance of reinforcement learning and supervised learning in this case, first, only segments of the test session data where measured RTs exists are considered. We calculate the rooted mean square errors (RMSE) between the measured RT and the predicted RT for the reinforcement learning model and the supervised learning model respectively. Second, to investigate the overall consistency between the measured RT and predicted RT, we interpolate all segments of the test data based on the measured RT using spline functions and calculate the correlation with the predicted RTs for each model. Fig. 9 illustrates the test cases for involved subjects, and TABLE II reports the corresponding statistics. In TABLE II, the first row shows the identifiers of the investigated subjects. Rows of RMSE indicate the average difference between the measured RT and predicted RT (in the units of seconds) via different methods for each subject. Rows of correlation show the correlation coefficients that reflect the coherence between the measured RT and predicted RT in a global trend.

TABLE II  RMSE AND CORRELATIONS OF RL AND SL FOR THE SINGLE-SUBJECT CASE

| Subjects | - | S01 | S09 | S41 |
|---|---|---|---|---|
| **RMSE** | SL | 1.26 | 1.68 | **1.87** |
|  | RL | **1.19** | **1.50** | 2.29 |
| **Correlation** | SL | 0.50 | 0.26 | 0.20 |
|  | RL | **0.62** | **0.55** | **0.35** |

SL: supervised learning; RL: reinforcement learning

From TABLE II, it is manifest that with almost the same network architectures but different learning paradigms, reinforcement learning potentially outperforms supervised learning in most cases. For S01 and S09, reinforcement learning has lower RMSEs but higher correlations. For S41, although the RMSE for reinforcement learning is not as good as supervised learning, the former still obtains a higher correlation coefficient.

The underperformance of the reinforcement learning model on S41 leads us to further investigate the characteristics of the training set to comprehend this behavior. The RT distributions for the data of two sessions involve in training are shown in Fig 10.

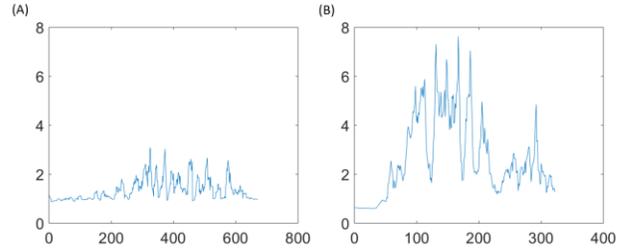

Fig. 10  RT distributions of two sessions (A) and (B). The $x$-axis indicates the sequence of RT in the session. The $y$-axis indicates the length of RT (unit: seconds).

It is observed that the RT distribution of the session in Fig. 10 (A) is eminently different from that in Fig. 10 (B). Most RT values in Fig. 10 (A) are less than 2 s. It is well known that one issue for Q-learning is the preference for overestimated values over underestimated values, especially when deterministic policy is adopted [38, 45]. During the training process, high chances of encountering small RTs will certainly drive the network to prefer small RT estimations, and this behavior will unavoidably migrate to test stage. That explains the discrepancy between the measured RTs and predicted RTs, as observed from the bottom-right image of Fig. 9. This phenomenon encourages us to consider additional tactics to mitigate the tendency of discrepancy for reinforcement learning as perspective.

### B. Cross-subject Case

For cross-subject performance assessment, we consider subjects that meet the same requirements as those for the single-subject case but participated in the experiment one or two times. The subjects that satisfy these criteria are S48(1)(350), S49(2)(705), S50(1)(361), S52(1)(239), and S53(2)(754). The numbers in the parentheses have the same meanings as those in the single-subject case. We adopt a leave-out testing paradigm here. In detail, we aggregate one session data from all subjects except for a specific subject for training, and use the session data from the specific subject for testing. The session data for validation are chosen only from the subjects that participated in the experiments two times, and one of these two session data are aggregated for validation.

Usually, the validation set shares an identical statistical distribution as the training set, although this property cannot be strictly satisfied here. This convention encourages us to use data from S49 and S53 for training and validation only. For convenience, we use data from S48, S50 and S53 for testing. For example, if the one session data of S48 is used for testing, then one session each from S49, S50, S52 and S53 is collected for training. The remaining one session each from S49 and S53 are collected for validation, typically to decide the number of training iterations.

Furthermore, besides the basic pre-processing of the EEG data as in the previous experiment, no normalization or other manipulations are adopted to mitigate the impact of cross-subject variance. To reduce the potential risk of overfitting and to guarantee good generalization, we use the double Q-network to analyze the data from multiple subjects, because compared with dueling Q-network, double Q-network has fewer

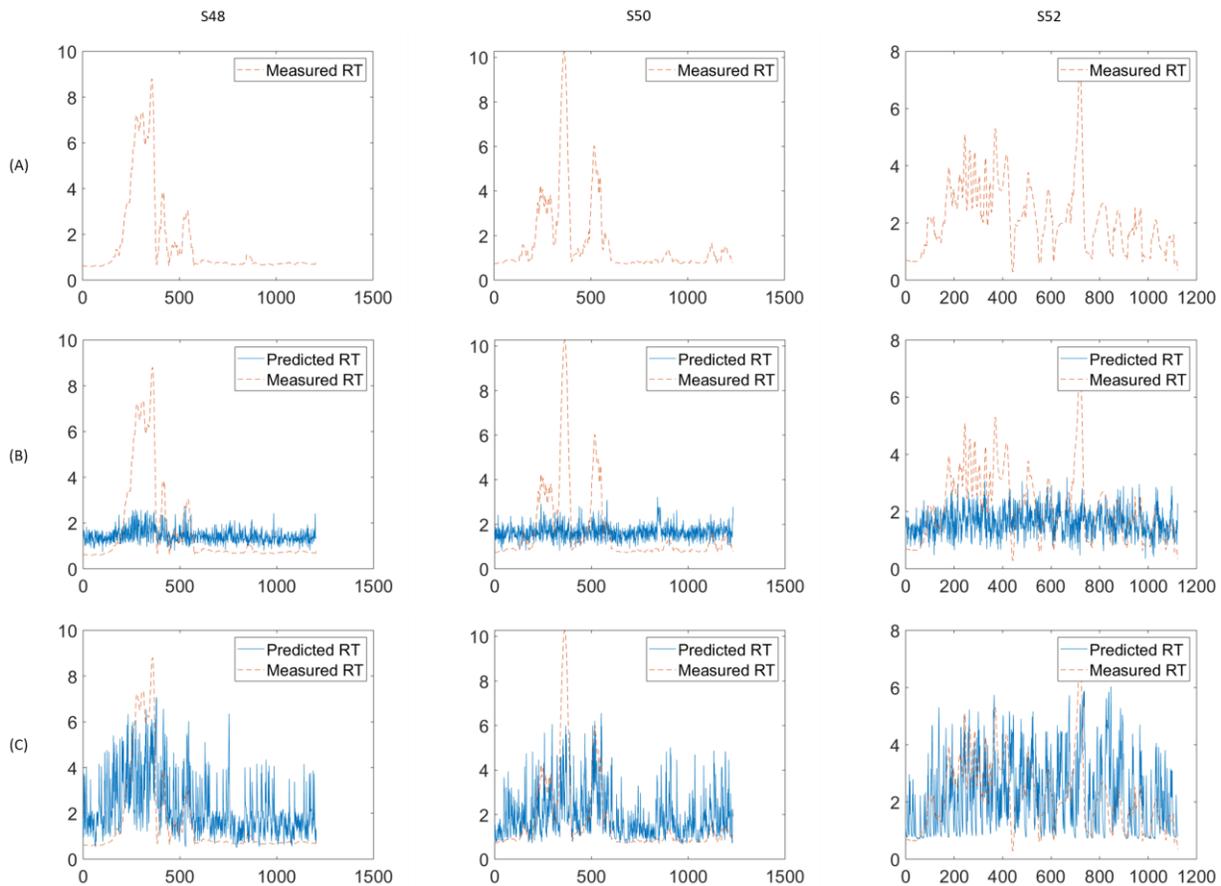

Fig. 11 Measured RT vs predicted RT for the cross-subject test: (A) measured RT; (B) supervised learning case; (C) reinforcement learning case. The $x$-axis indicates the progress of the session (3 seconds per step). The $y$-axis indicates lengh of RT (unit: seconds).

parameters.

Following similar procedures and adopting the same hyperparameters as those for the single-subject case (for subject S50, we reduce the transition weight from 0.75 to 0.6), TABLE III presents the resulting statistics for both supervised learning and reinforcement learning, and Fig. 11 shows the predicted and measured RT of the test data.

The meanings of the fields in TABLE III are the same as in TABLE II. From TABLE III, it can be observed that the RMSE of the prediction of supervised learning is better than the reinforcement learning counterpart. However, as previously mentioned, the discrepancy of distributions between training data and test data makes the generalization a challenging task for supervised learning. Reinforcement learning is more robust in this regard, as shown by the correlation coefficients. It indicates that the model tries to trace the alteration of RT in an overall effective manner, which is more significant from an application perspective.

The above results, especially the high correlation coefficients, demonstrate the effectiveness of introducing DQN for EEG data analysis from the session level. If calibration is not a problem, our method can be a suitable candidate for BCI application especially for the single-subject case. In addition, we postulate that if the experiment is redesigned from the multiple-session perspective to have more session data to train the network, the results can be further enhanced.

One trait of the predicted RT of reinforcement learning in Fig. 9 and Fig. 11 is the high variance of the prediction, which is more obvious than the supervised learning counterpart. This is due to a known effect of the deterministic policy gradient [36], which is a limited case of the stochastic policy gradient, providing the variance parameter $\sigma = 0$. However, $\sigma$ greater than zero is quite common in practice, which leads to high variance when approximating the optimal policy. How to reduce the variance is another direction to improve the performance of our proposed model in future work.

## V. DISCUSSION

In this discussion, we first discuss some background of this research and then some technical details regarding the experiment.

First, we elaborate on the initiation of this research. One notorious problem of EEG research, especially from the cognitive perspective via machine learning, is the label problem. It is understandable that current limited achievements

TABLE III RMSE AND CORRELATIONS OF RL AND SL FOR CROSS-SUBJECT CASE

| Subjects | - | S48 | S50 | S52 |
|---|---|---|---|---|
| **RMSE** | SL | **1.31** | **1.10** | **1.19** |
|  | RL | 1.57 | 1.16 | 1.21 |
| **Correlation** | SL | 0.39 | 0.07 | 0.16 |
|  | RL | **0.48** | **0.45** | **0.33** |

SL: supervised learning; RL: reinforcement learning of neuroscience still prohibit direct probing of the mind state, such as the degree of drowsiness during driving. Indirect methods, such as RT in this paper, mitigate this difficulty; however, some bias is unavoidably introduced. In contrast, considering image recognition, there is usually no ambiguity regarding what is contained in the image. We can identify an image of either a dog or a cat, but not something in between. However, due to factors such as a sudden drift of mind or different muscle motor trajectories, the measured RT is ineluctably contaminated with noise, but these RTs are still presented as ground-truth labels. Retrospect of our research conducted during past years helps realize that following supervised learning paradigm to train a complicated model solely for predicting RT chasing SOTA accuracy is doomed to poor generalization during deployment. A promising direction is to harness the available but inaccurate RT to globally trace the procedural change of the mind state. As demonstrated in our paper, the working principle of reinforcement learning coincides with such a special requirement for safety driving monitoring via EEG. Because, during the optimization of reinforcement learning, the reward of the current action can be seen as an extrapolation of future measured RT into the current step, and it does not have to be accurate. However, the goal-directed approach can fit the predicted RT to follow the trend. This approach relieves the high-quality requirement on ground-truth labels but retain the drive to trace the overall mind state change.

Currently we only explore Q-learning, specifically DQN, to address the problem. First, it can be noted in Fig. 4(A) that the action which is issued to regulate the internal RT tracer has no effect on the mind state. Alternatively speaking, the transition from state $s_t$ to $s_{t+1}$ is unconditioned on $a_t$. Because model-based RL tries to predict the next state transition based on the previous state and action, the problem formulation is not feasible for model-based RL. For model-free RL, the solutions include tabular methods, represented by Q-learning, and approximating methods, represented by actor-critic, for example. The latter is not suitable here due to dimensional constraints. Considering the navigation problem in [46], based on the raw input image, the state can be abstracted to position and head direction. These latent representations of state can be easily combined with limited action choices in a balanced manner and consequently fed into the critic network. However, for EEG data, there lacks such a meaningful abstraction. The feature dimension is still quite large compared with the number of choices, which is impractical to concatenate the state and action together. This constraint reduces the problem to only a DQN solution. Investigation of other RL methods is planned as one of our future works.

Next, we unveil and discuss some interesting technical details concerning the experiment. Although research has clearly revealed the positive correlation between the mind state and RT [13, 14], there is no explicit formula to express the relation between the two. Even if this formula exists, due to the current unmeasurable mind state, the correctness of such a formula would be under dispute. However, if some findings in previous work could be consolidated with the research in this paper, the outcomes would be interesting and beneficial.

TABLE IV RMSE AND CORRELATIONS FOR DIFFERENT TRANSITION WEIGHT VALUES

| $\beta$ | 0.2 | 0.4 | 0.6 | 0.75 | 0.8 |
|---|---|---|---|---|---|
| **RMSE** | 1.63 | 1.28 | 1.20 | 1.19 | 1.15 |
| **Correlation** | 0.40 | 0.61 | 0.62 | 0.62 | 0.60 |

In [47], the authors indicated that the distinguishable change in mind state will be beyond 4 minutes. Referring to the transition weight $\beta$ in (3), to catch the alteration of RT, our initial setting of $\beta$ is 0.2, to merit the proposed RT based on the current state (EEG data). As revealed in TABLE IV, poor performance seems indicate a false impression that reinforcement learning might not work. However, a review of the related research urges us to try greater $\beta$ values. The current network structure indicates that $\beta = 0.75$ is a preferred value, which is a good balance between the RMSE and correlation coefficient, even if we are not sure it is the optimal value when we model the mind state transition as in (3).

Any value of $\beta$ greater than 0.5 indicates that the network favours the traced RT that is historically maintained over the newly proposed RT. Although the measured RT tends to be problematic and noisy, it is still a good indicator of the procedural mind state alterations. Aligning RTs with EEG data via our model consolidates the conclusion that the mind state changes in a procedural way [13, 48], which can guide the design of a better apparatus for driving safety. For example, if the current mind state is regarded as alert, a sudden prolonged predicted RT should not incur some warning, because it tends to be a false alarm.

## VI. CONCLUSION

In this paper, we introduced deep reinforcement learning specifically deep Q-learning for fatigue estimation to target driving safety. We constructed variants of the deep Q-network and used them to carry out the experiment for evaluating our methodology. The results manifested the practicality via high correlation coefficients between the measured RT and predicted RT in both the single-subject case and cross-subject case. Due to reinforcement learning's low dependency on the quality of the label information and the high efficiency of data utilization, our work calls for potential future research to systematically consider reinforcement learning in BCI for different applications.

ACKNOWLEDGEMENT

This work was supported in part by the Australian Research Council (ARC) under discovery grant DP180100670 and DP180100656; NSW Defense Innovation Network and NSW State Government of Australia under the grant DINPP2019 S1-03/09; Office of Naval Research Global, US under Cooperative Agreement Number ONRG-NICOP-N62909-19-1-2058.